%% file: CVPR-2022-AMT-GAN.tex
\documentclass[10pt,twocolumn,letterpaper]{article}

\usepackage[pagenumbers]{cvpr} 

\usepackage{graphicx}
\usepackage{amsmath}
\usepackage{amssymb}
\usepackage{booktabs}
\usepackage{amsfonts,amssymb}
\usepackage{algorithm}
\usepackage{algorithmic}
\usepackage{multirow}
\usepackage{float}
\usepackage{caption}
\usepackage{subcaption}
\usepackage{fontawesome}
\usepackage{xcolor}
\usepackage{colortbl}
\usepackage{url}
\usepackage[accsupp]{axessibility}

\newcommand{\black}[1]{\textcolor{black}{#1}}

\def\ie{\textit{i.e.}}

\usepackage[pagebackref,breaklinks,colorlinks]{hyperref}

\usepackage[capitalize]{cleveref}
\crefname{section}{Sec.}{Secs.}
\Crefname{section}{Section}{Sections}
\Crefname{table}{Table}{Tables}
\crefname{table}{Tab.}{Tabs.}

\def\cvprPaperID{10243} 
\def\confName{CVPR}
\def\confYear{2022}

\begin{document}

\title{\black{Protecting Facial Privacy: Generating Adversarial Identity Masks via Style-robust Makeup Transfer}}

\author{Shengshan Hu\textsuperscript{\rm 1, 4, 5, 6}, Xiaogeng Liu\textsuperscript{\rm 1, 4, 5, 6}, Yechao Zhang\textsuperscript{\rm 1, 4, 5, 6},  Minghui Li\textsuperscript{\rm 3}\\ Leo Yu Zhang\textsuperscript{\rm 8}, Hai Jin\textsuperscript{\rm 2, 4, 5, 7}, Libing Wu\textsuperscript{\rm 9}\\
\textsuperscript{\rm 1}School of Cyber Science and Engineering, Huazhong University of Science and Technology\\
\textsuperscript{\rm 2}School of Computer Science and Technology, Huazhong University of Science and Technology\\
\textsuperscript{\rm 3}School of Software Engineering, Huazhong University of Science and Technology\\
\textsuperscript{\rm 4}National Engineering Research Center for Big Data Technology and System\\
\textsuperscript{\rm 5}Services Computing Technology and System Lab\\
\textsuperscript{\rm 6}Hubei Engineering Research Center on Big Data Security\\
\textsuperscript{\rm 7}Cluster and Grid Computing Lab\quad
\textsuperscript{\rm 8}School of Information Technology, Deakin University \\
\textsuperscript{\rm 9}School of Cyber Science and Engineering, Wuhan University\\
{\tt\small \{hushengshan, liuxiaogeng, ycz, minghuili, hjin\}@hust.edu.cn}\\ 
{\tt\small leo.zhang@deakin.edu.au, wu@whu.edu.cn}
}

\maketitle

\begin{abstract}
While deep face recognition (FR) systems have shown amazing performance in identification and verification, they also arouse privacy concerns for their excessive surveillance on users, especially for public face images widely spread on social networks. Recently, some studies adopt adversarial examples to protect photos from being identified by unauthorized face recognition systems. However, existing methods of generating adversarial face images suffer from many limitations, such as awkward visual, white-box setting, weak transferability, making them difficult to be applied to protect face privacy in reality.

In this paper, we propose adversarial makeup transfer GAN (AMT-GAN)\footnote{\url{https://github.com/CGCL-codes/AMT-GAN}}, a novel face protection method aiming at constructing adversarial face images that preserve stronger black-box transferability and better visual quality simultaneously. AMT-GAN leverages generative adversarial networks (GAN) to synthesize adversarial face images with makeup transferred from reference images. In particular, we introduce a  new regularization module along with a joint training strategy to reconcile the conflicts between the adversarial noises and the cycle consistence loss in makeup transfer, achieving a desirable balance between the attack strength and visual changes. Extensive experiments verify that compared with state of the arts, AMT-GAN can not only preserve a comfortable visual quality, but also achieve a higher attack success rate over commercial FR APIs, including Face++, Aliyun, and Microsoft.
\end{abstract}

\input{section/1-Introduction}

\input{section/2-Related_Works}

\input{section/3-method_ver2.0}

\input{section/4-experiments}
\input{section/6-Limitation}

\input{section/5-conclusion}

~\\
\noindent\textbf{Acknowledgments.} Shengshan’s work is supported in part by the National Natural Science Foundation of China (Grant Nos. 62002126, U20A20177), and Fundamental Research Funds for the Central Universities (Grant No. 5003129001). Leo’s work is supported in part by the National Natural Science Foundation of China (Grant No. 61702221). Libing's work is supported by Key R\&D plan of Hubei Province (No. 2021BAA025). Minghui Li is the corresponding author.

{\small
\bibliographystyle{ieee_fullname}
\bibliography{egbib}
}

\input{section/Appendix}
\end{document}

%% file: section/1-Introduction.tex
\section{Introduction}\label{intro}

\begin{figure}[t]
\setlength{\belowcaptionskip}{-0.5cm}
\centering
\includegraphics[width=0.35\textwidth]{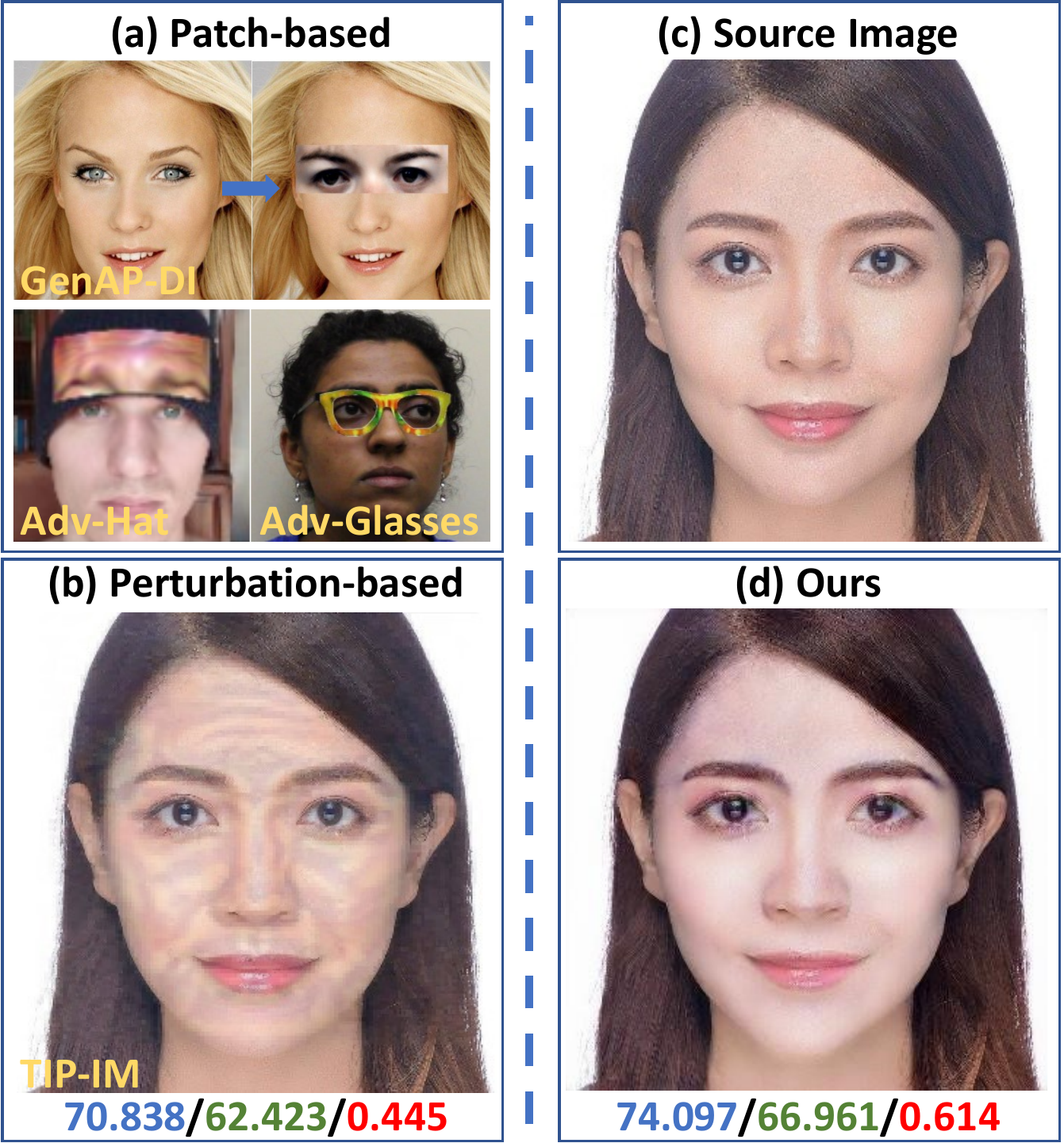}
\caption{Comparison with existing adversarial attacks on FR systems in the black-box setting. The images in (a) are directly extracted from their papers. The numbers listed below images are the verification confidence of the target identity given by commercial FR APIs, and a higher score represent a stronger attack ability. The blue is from Face++, the green is from Aliyun, and the red is from Microsoft Azure {(The same color code will be used hereinafter)}.}
\label{fig:introduction}
\end{figure}

Recent years have witnessed the fast development of \textit{face recognition} (FR) based on \textit{deep neural networks} (DNNs). The powerful face recognition systems, however, also pose a great threat to personal privacy. For example, it has been shown that FR systems can be used to identify social media profiles and track user relationships through large-scale photo analysis~\cite{DBLP:journals/popets/Shoshitaishvili15,hill2020secretive}. Such kind of excessive surveillance on users urgently demands an effective approach to help individuals protect their face images against unauthorized FR systems. 

\black{Launching \textit{data poisoning attacks} towards the \black{training dataset or the gallery dataset }of malicious FR models is a promising solution to protect facial privacy~\cite{shan2020fawkes,DBLP:conf/iclr/CherepanovaGFDD21}. However, these schemes require that the adversary (\ie, the users in our scenario) can inject poisoned face images into the \black{datasets}. Once the target model is trained or makes inference over clean datasets, they are likely to become invalid.} 

Another strategy is to make use of \textit{adversarial examples} to \black{launch \textit{evasion attacks} and }protect face images from being illegally identified~\cite{DBLP:journals/popets/RajabiBRWF21,DBLP:journals/corr/SzegedyZSBEGF13,yang2021towards}. Adversarial face images are more suitable for protecting user privacy in real-world scenarios since they only need to modify users' own data regardless of the settings of the target model. Unfortunately, existing approaches of generating adversarial face examples suffer from several limitations when they are considered to protect face privacy on social media: (1) \textbf{Accessibility to target models.} Most existing schemes belong to white-box attack (\ie, the adversary has full knowledge of the target model)~\cite{DBLP:journals/corr/GoodfellowSS14,DBLP:conf/iclr/MadryMSTV18}, or query-based black-box attack (\ie, the adversary can arbitrarily query the target model)~\cite{DBLP:conf/cvpr/DongSWLL0019}. They are infeasible for protecting users' privacy since the users have no idea which kind of DNNs the third-party tracker is running; (2) \textbf{Poor visual quality.} As shown in Fig.~\ref{fig:introduction}, existing adversarial attacks on FR system fail to preserve the image quality in the black-box setting. Patch-based adversarial attacks~\cite{DBLP:conf/icpr/KomkovP20,DBLP:conf/cvpr/XiaoGFDGZ0021} often cause a fairly bizarre and conspicuous change on source images, and the state-of-the-art perturbation-based method~\cite{yang2021towards} makes the modified face fill with awkward noises; (3) \textbf{Weak transferability.} Fig.~\ref{fig:introduction} also demonstrates that the state of the art has a relatively low attack success rate on commercial APIs. \textit{In summary, it is still challenging to balance the trade-off between the visual quality and attack ability of adversarial face images in the black-box setting.}

 
In this paper, \black{we  solve this problem from  a new perspective.}
\black{Different from existing works trying to place multifarious restrictions on perturbations and then dig a better gradient-based algorithm to construct adversarial examples,  we focus on organizing the perturbations, although extensive and visible, in a reasonable way such that they appear natural and comfortable, under the condition that a high attack ability is maintained. We therefore leverage makeup as the key idea for arranging perturbations. Specifically, }we propose a new framework called \textit{adversarial makeup transfer GAN} (AMT-GAN) to generate adversarial face images with the natural appearance and stronger black-box attack strength. ATM-GAN first exploits a set of generative adversarial networks to construct adversarial examples that can inherit makeup styles from a reference image. In order to reconcile the conflicts between the adversarial noises and the cycle consistence loss in makeup transfer, we incorporate a 
newly designed regularization module by exploiting the disentanglement function of the encoder-decoder architecture and the residual dense blocks in the image super-resolution. As a result, the adversarial toxicity can be compatibly alleviated in the cycle reconstruction phase, making the generator focus on building robust mappings between the source domain and the target style domain with adversarial features. \black{In addition, we introduce a joint training strategy  which integrates the traditional \textit{G-D} game in the GAN training and the newly designed regularization module, as well as the transferability enhancement process to encourage the generator to catch, imitate, and reconstruct the common adversarial features which can effectively transfer between different models.}

\black{To the best of our knowledge, we propose the first \black{joint training framework} to address the collapse phenomenon of the cycle consistency and the domain mappings of the generator when the image-to-image translation GANs are used to craft adversarial examples. Our joint training framework can be extended to other security-sensitive fields when GAN is considered, such as Deepfake~\cite{westerlund2019emergence}. }In summary, we make the following contributions:
\begin{itemize}
\item We propose AMT-GAN, a more practical approach for protecting face images against unauthorized FR systems, by constructing adversarial examples with outstanding black-box attack performance and natural appearance that derive cosmetic styles from any chosen reference images.
\black{\item We design a regularization module based on feature disentanglement to improve the visual quality of adversarial images, and then develop a joint training pipeline to train the generator, the discriminator, and the regularization module, such that the generator can accomplish two  jobs (\ie, makeup-transfer and adversarial attack) simultaneously and build robust mappings among different data manifolds.}
\item Our extensive experiments on multiple benchmark datasets verify that AMT-GAN is highly effective at attacking various deep FR models, \black{including }commercial face verification APIs such as Face++ \footnote{\url{https://www.faceplusplus.com}}, Aliyun\footnote{\url{https://vision.aliyun.com}}, and Microsoft\footnote{\url{https://azure.microsoft.com}}, where we outperform state of the arts about $4\%\sim60\%$. 
\end{itemize}

%% file: section/2-Related_Works.tex
\begin{figure*}[t]
\setlength{\belowcaptionskip}{-0.1cm}
\centering
\includegraphics[width=0.95\textwidth]{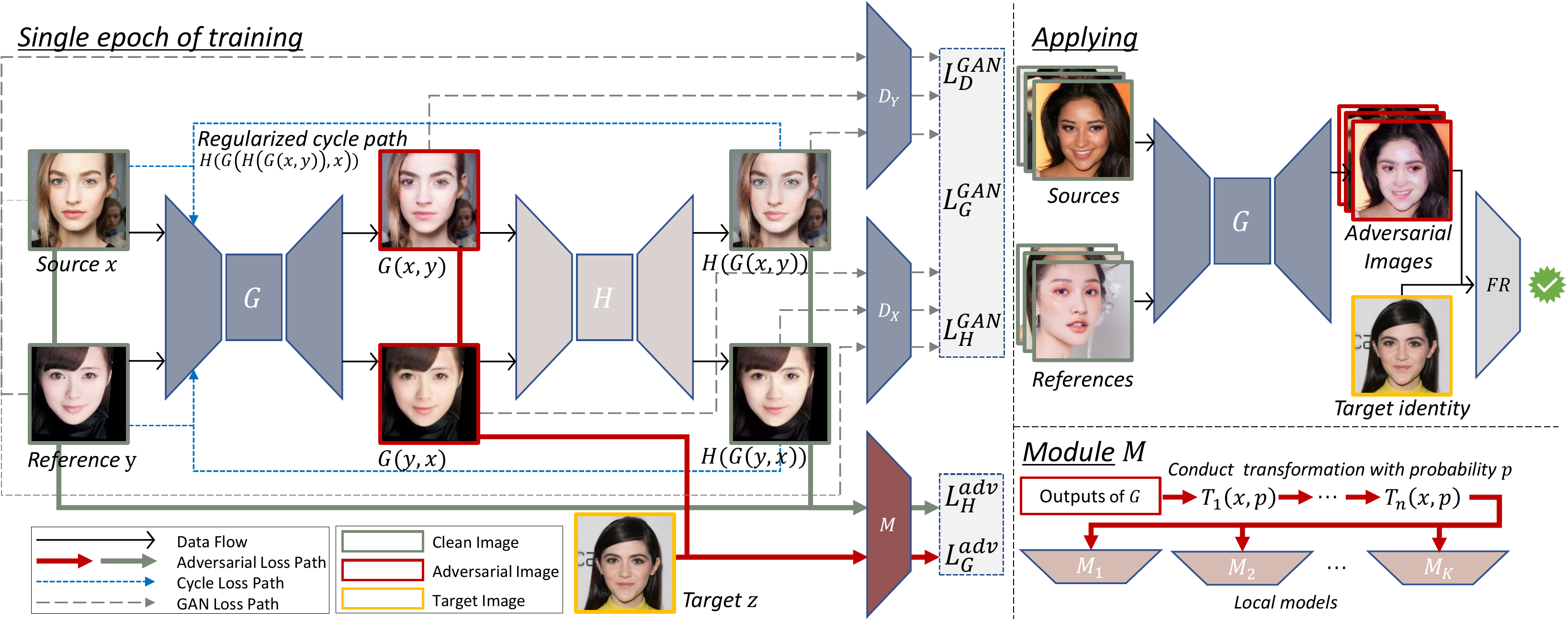}
\caption{The architecture of AMT-GAN}
\label{fig:pipe}
\end{figure*}
\section{Related Works}\label{relatedwork}
To protect face privacy, AMT-GAN leverages adversarial examples which are originally designed to disable DNNs in the machine learning community. Thus, we first discuss existing works on adversarial attacks in the context of privacy protection, followed by recent research on style transfer.
\subsection{Adversarial Attacks on Face Recognition}
Broad studies have shown that the DNNs are fatally vulnerable to adversarial examples~\cite{DBLP:journals/corr/SzegedyZSBEGF13,DBLP:journals/corr/GoodfellowSS14,DBLP:conf/iclr/MadryMSTV18}, and many adversarial algorithms have been developed to attack face recognition system~\cite{DBLP:conf/cvpr/DongSWLL0019,DBLP:journals/corr/abs-1801-00349,DBLP:conf/eccv/QiuXYYLL20,DBLP:conf/cvpr/XiaoGFDGZ0021}. Depending on the adversary’s knowledge of the target model, the adversarial face attacks can be divided into white-box attacks and black-box attacks.

The white-box attacks~\cite{DBLP:journals/corr/GoodfellowSS14,DBLP:conf/iclr/MadryMSTV18,hu2021advhash} and query-based black-box attacks~\cite{DBLP:conf/cvpr/DongSWLL0019,DBLP:journals/corr/abs-2104-06728} heavily rely on the accessibility to the target model, which is a stringent prerequisite in practice. 
Transferability-based black-box attacks are thus more suitable for protecting face images in real-world scenarios~\cite{DBLP:conf/cvpr/DongLPS0HL18,DBLP:conf/cvpr/XiaoGFDGZ0021,DBLP:journals/tifs/ZhongD21,yang2021towards}.
However, most existing transferability-based attacks~\cite{DBLP:conf/cvpr/DongLPS0HL18,DBLP:conf/cvpr/XieZZBWRY19} are designed to solve an optimization problem, which is not only time-consuming, but also likely to be trapped in over-fitting and degrades the transferability~\cite{DBLP:conf/cvpr/XiaoGFDGZ0021}. In addition, to maintain the attack strength on different black-box models, transferability-based attacks usually generate images with perceptible noises~\cite{DBLP:conf/cvpr/DongLPS0HL18,DBLP:conf/cvpr/DongPSZ19}. Recent work~\cite{yang2021towards} tried to solve this problem by adding a new penalty function to fit the privacy-preserving scenario. However, as illustrated in Fig.~\ref{fig:introduction}, the perturbations are still outrageous. Using GAN~\cite{DBLP:journals/corr/GoodfellowPMXWOCB14} to craft adversarial samples~\cite{DBLP:conf/ijcai/XiaoLZHLS18} can achieve improvement in terms of image noises, but it is still challenging to maintain GAN's stability for data with high-definition in adversarial task.~\cite{DBLP:conf/eccv/QiuXYYLL20} turns to craft adversarial samples by swapping the latent features of face images from a GAN, however, it changes original face attributes dramatically (such as turning closed mouth to open), which are not desirable for users in social media. The patch-based adversarial attacks on FR systems~\cite{DBLP:journals/corr/abs-1801-00349,DBLP:conf/icpr/KomkovP20,DBLP:conf/ijcai/YinWYGKDLL21,DBLP:conf/cvpr/XiaoGFDGZ0021} are also not compatible for the same reason.

\subsection{Style Transfer and Makeup Transfer}
Style transfer~\cite{DBLP:conf/cvpr/ChoiCKH0C18} is an image-to-image translation technique that aims to separate and recombine the content and style information of images. Built on the style transfer framework, makeup transfer~\cite{DBLP:conf/cvpr/ChangLYF18,DBLP:conf/mm/LiQDLYZL18,DBLP:conf/iccv/GuWCTT19,DBLP:conf/cvpr/JiangLG0HFY20,Deng_2021_CVPR} is proposed to transfer the makeup style of the reference image to the source image while keeping the result of face recognition unchanged. Both style transfer and makeup transfer rely on the cycle consistency loss or its variants~\cite{DBLP:conf/iccv/ZhuPIE17,DBLP:conf/eccv/ZhaoWD20} to maintain the stability of source images. 

Recently, \black{\cite{DBLP:conf/icip/ZhuLC19} made the first attempt to exploit the makeup transfer to generate adversarial face images in a white-box setting. Then \cite{DBLP:conf/ijcai/YinWYGKDLL21} tried to construct adversarial cosmetic face images with transferability property to realize black-box attacks. However, it not only has a low attack success rate, but also fails to preserve the image visual quality where the modifications added to the source image is abnormal and noticeable, especially when the styles between reference and source images are significantly different, as explicitly depicted in our experiments Fig.~\ref{fig.compare_ijcai}.}

\black{It is indicated that adversarial noise may cause dysfunction on the cycle consistency loss~\cite{10.1007/978-3-030-66823-5_14,9533868}, and gets confirmed in this paper. In the literature, it is still challenging to amicably incorporate makeup transfer into adversarial examples, generating a natural adversarial face image that maintains a high attack success rate on black-box face recognition systems.}



%% file: section/3-method_ver2.0.tex
\section{Adversarial Makeup Transfer GAN (AMT-GAN)}\label{method}
\subsection{Problem Formulation}
In this section, we formulate the problem of adversarial attacks with makeup transfer on FR systems. To protect facial privacy effectively against malicious FR models, we mainly consider targeted adversarial attack (\ie impersonation attack) that aims to generate adversarial examples which can be recognized as the specified target identity. Generally, the targeted adversarial attack on FR system can be formulated as:
\begin{equation}\label{eq:adversarial fomulation}
        \min \limits_{x^A} L_{adv} = D(M_k(x^A),M_k(z)),
\end{equation}
where $D(\cdot)$ represents a distance function such as cross-entropy or cosine similarity, $M_k$ represents a DNN-based feature extractor for FR, $x^A$ and $z$ stand for the adversarial face image and the target image respectively.


As for makeup transfer, let \textit{$X, Y \subset \mathbb{R}^{H \times W \times 3}$} denote the makeup style domain of the source and reference images, respectively. Here, we use $x \in X$ and $y \in Y$ to represent the clean face images and $x^A \in X$ and $y^A \in Y$ to represent their adversarial face images, respectively. The adversarial makeup transfer is expected to train a function \textit{$G$}: $\{x, y\} \to \tilde{y}^A_x$, where the adversarial image \textit{$\tilde{y}^A_x$} has the same makeup style with \textit{$y$} and the same visual identity with \textit{$x$}.

\begin{figure}[t]
\setlength{\belowcaptionskip}{-0.1cm}
    \begin{subfigure}{0.22\textwidth}
        \includegraphics[width=\textwidth]{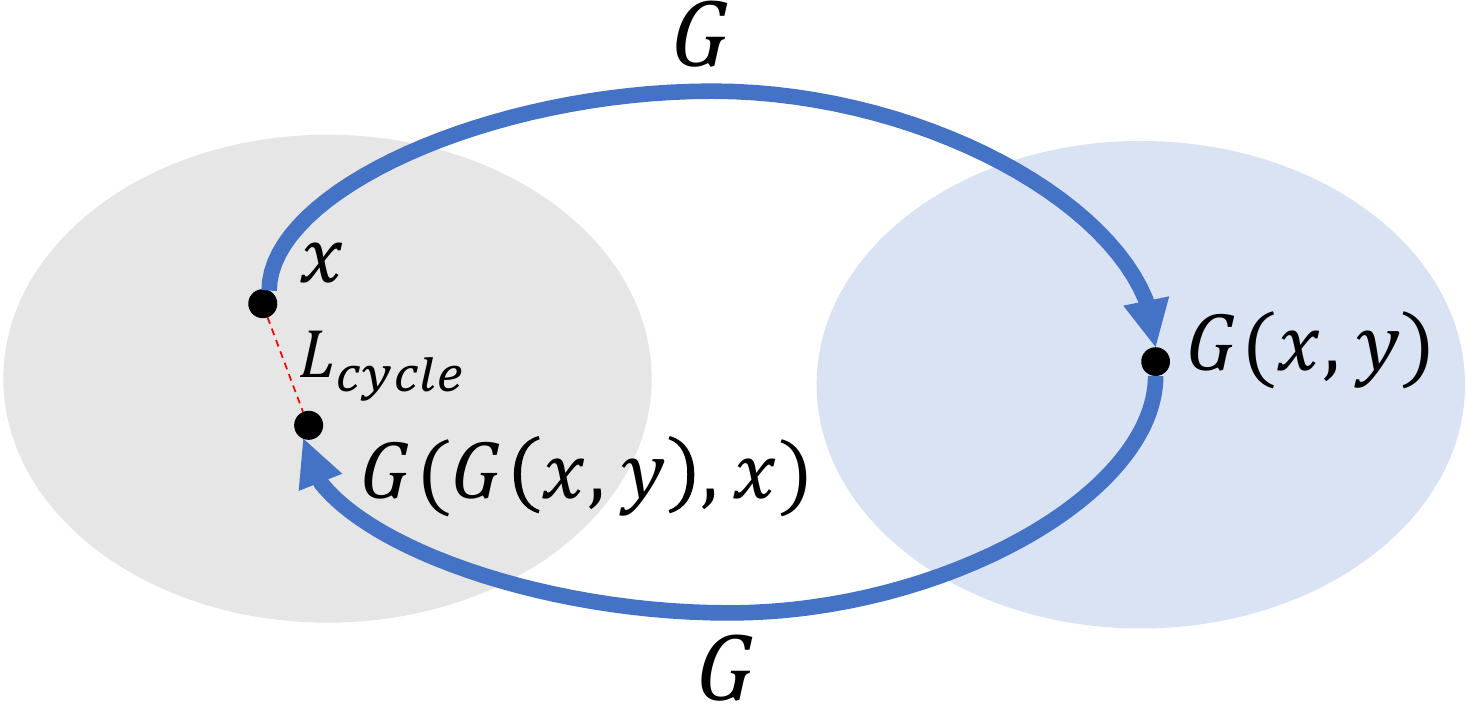} 
        \caption{}
        \label{Fig.org_cycle}
    \end{subfigure}
    \hfill
    \begin{subfigure}{0.22\textwidth}
        \includegraphics[width=\textwidth]{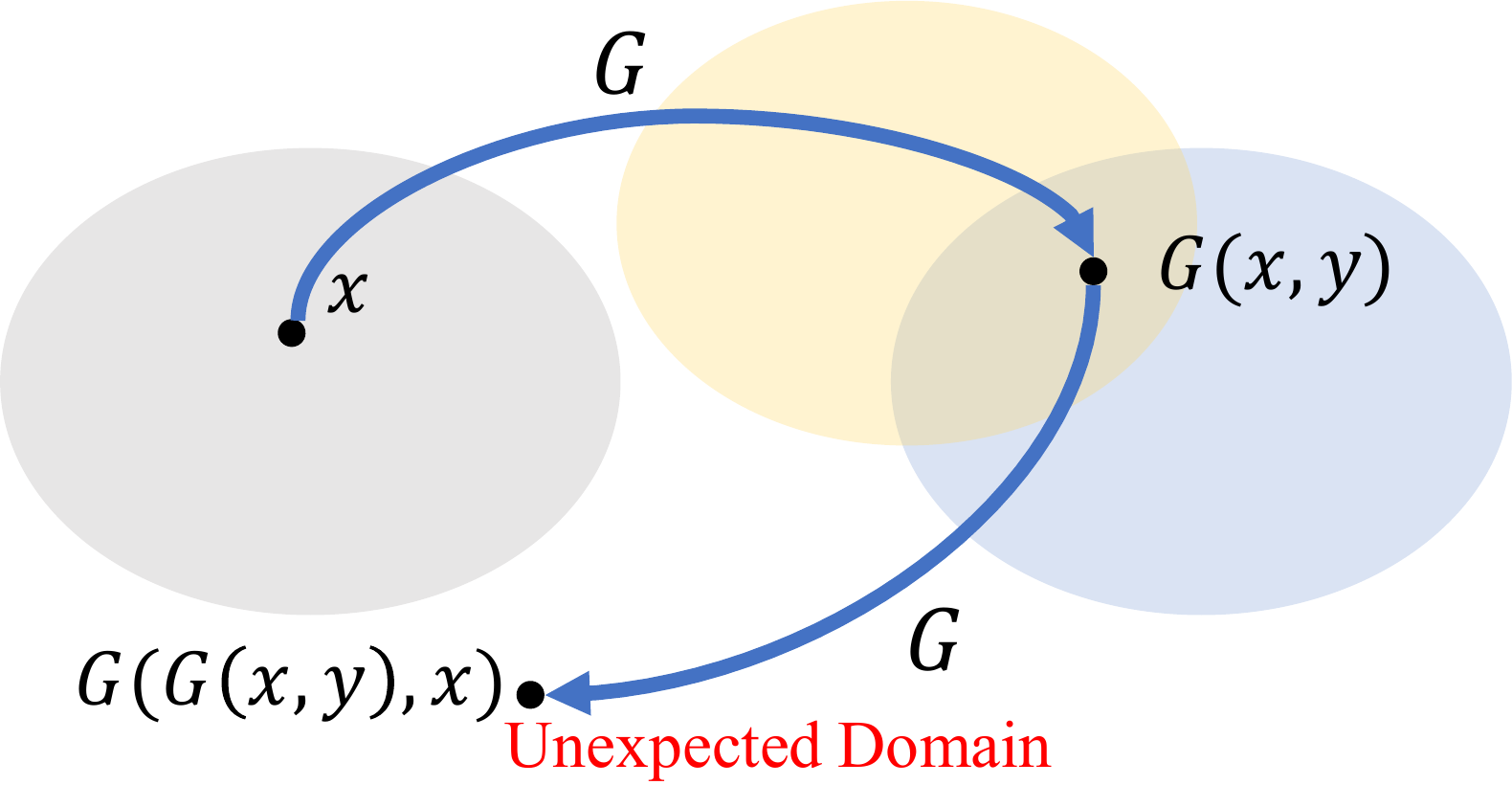} 
        \caption{}
        \label{Fig.fail_cycle}
    \end{subfigure}
    \begin{subfigure}{0.48\textwidth}
        \includegraphics[width=\textwidth]{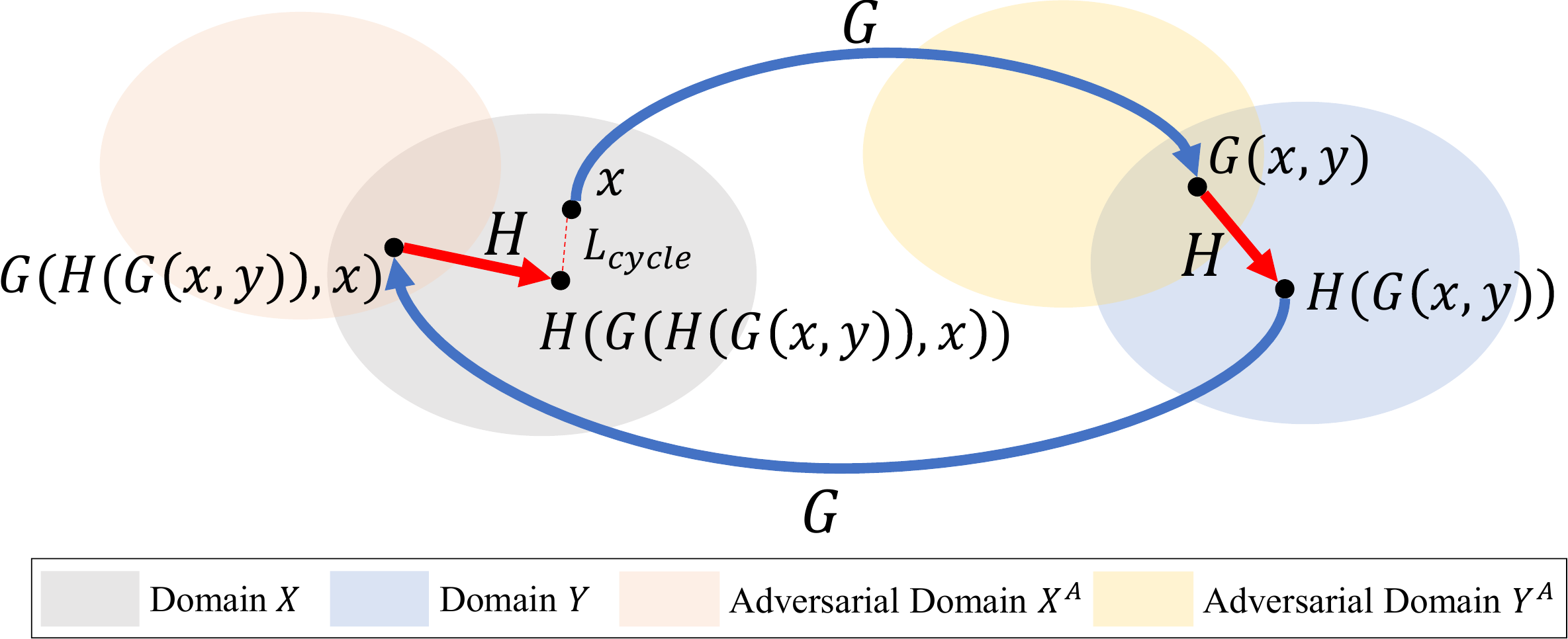} 
        \caption{}
        \label{Fig.H_cycle}
    \end{subfigure}
    \caption{ (a) Normal loop path of cycle consistency loss; (b) Damaged loop path caused by adversarial \black{noises}; (c) \black{Recovered} loop path of our regularized cycle consistency loss.}
\end{figure}

\subsection{Detailed Construction}
The architecture of AMT-GAN is depicted in Fig~\ref{fig:pipe}.

\textbf{ Generator \textit{$G$} and discriminators $D_X,D_Y$.} 
The generator $G$ is supposed to generate adversarial examples for source images, while ensuring that the visual identity remains the same but the makeup style changes from the source domain to the reference domain.
The discriminators $D_X$ and $D_Y$ are supposed to distinguish the distribution of fake images generated by $G$ from that of the real images. 
Mathematically, the loss functions of GANs are formulated as:



\begin{equation}\label{eq:LossGAN_D}
\begin{aligned}
    L^{gan}_D =&- \log D_X(x) - \log (1-D_X(G(y,x)))\\
    &- \log D_Y(y) - \log (1-D_Y(G(x,y))), 
\end{aligned}
\end{equation}
\begin{equation}\label{eq:LossGAN_G}
\begin{aligned}
    L^{gan}_G =&- \log (D_X(G(y,x))) - \log (D_Y(G(x,y))).
\end{aligned}
\end{equation}


To learn the bidirectional mappings (\ie, $X \to Y, Y \to X$) between the source and reference domain without supervised training data, we also utilize the cycle consistency loss function~\cite{DBLP:conf/iccv/ZhuPIE17}, which is a crucial element in the unsupervised image-to-image translation tasks. Generally, the cycle consistency loss function is depicted in Fig~\ref{Fig.org_cycle} and denoted as:

\begin{equation}\label{eq:LossCycle}
\begin{aligned}
    L^{cycle} &= \parallel G(G(x,y),x)  -x\parallel_1,
\end{aligned}
\end{equation}
\black{where $\parallel \cdot \parallel_1$ represents the $L_1$ norm.}


However, the cycle consistency loss is in conflict with the adversarial example.
As shown in Fig~\ref{Fig.fail_cycle}, when establishing the inverse mapping $Y^A \to X$, the generated adversarial examples $G(x, y)$ is taken as the first input of $G$. Due to the adversarial modification on the samples from the clean domain $X$, $G$ may fail to extract the features of $G(x, y)$ and is thus unable to turn an adversarial input from adversarial domain $Y^A$ back to the clean domain $X$. As a result, the recovered $G(G(x,y),x)$ may be significantly different from the source image $x$, \ie, $G(G(x,y),x)$ is in an unexpected domain rather than the domain $X$. In short, due to the adversarial nature of $G$'s outputs, it is hard for $G$ to establish a robust inverse mapping $Y^A \to X$. This phenomenon makes the existing cycle consistency loss unsuitable for generating adversarial face images with makeup transfer (see our ablation study Fig.~\ref{fig.ablation_H}).




  

\textbf{Regularization module \textit{$H$}.} 
To ensure the cycle consistency works well, we introduce a regularization module $H$ in our framework. $H$ is designed to generate clean images $H(G(x,y))$ with the same content, style, and dimensions as $G(x,y)$ but without adversarial property. Namely, $H$ transforms the generated image $G(x,y)$ from the adversarial domain $Y^A$ to the clean domain $Y$. The combination of $G$ and $H$ maintains a new cycle consistency, \ie, the loop $X \to Y^A, Y^A \to Y$, then $Y \to X^A, X^A \to X$, as illustrated in Fig.~\ref{Fig.H_cycle}.

To find an effective $H$, we first choose a pair of encoder and decoder as the basic architecture of $H$. Then we leverage the \textit{residual-in-residual dense block} (RRDB) as a key block of $H$. RRDB is widely used in the field of image super-resolution~\cite{DBLP:conf/eccv/WangYWGLDQL18}. These blocks will maintain and recover the content and texture information of the input while extracting and discarding the adversarial perturbations at the same time.

In our design, the regularization module $H$ follows the same training process as the main networks $G$ and $D$. 
The newly designed regularized cycle consistency loss is formulated as:

\begin{equation}\label{eq:LossCycleH}
\begin{aligned}
    L^{reg}_G = & \parallel H(G(H(G(x,y)),x))  -x\parallel_1 \\
    &+ \parallel H(G(H(G(y,x)),y))  -y\parallel_1.
\end{aligned}
\end{equation}


In addition, $H$ is supposed to keep the output of $G$ visually unchanged while alleviating the adversarial effects. So we let $H$ accept the penalty by the returns from $D_X$ and $D_Y$, which encourages $H$ to reconstruct the real performance of $G$ in style transfer, and it is defined as:

\begin{equation}\label{eq:LossGAN_H}
\begin{aligned}
    L^{gan}_H =&- \log (D_X(H(G(y,x)))) - \log (D_Y(H(G(x,y)))).
\end{aligned}
\end{equation}


Note that in Eq.~(\ref{eq:LossGAN_D}), we do not include the performance of $H$ in $L^{gan}_D$, as  we want to keep the two-player zero-sum game between the generator and the discriminators stable, and it is also unnecessary to adjust $D_X$ and $D_Y$ according to the performance of $H$.

\textbf{\black{Transferability enhancement module \textit{$M$}.}} \textit{$M$} consists of \textit{$K$} pre-trained face recognition models $\{M_k\}_{k=1,...,K}$, which have high accuracy on public face images datasets. These local models serve as white-box models when we train our GANs and try to imitate the decision boundaries of potential target models which we cannot access. 

 




In our method, inspired by~\cite{DBLP:conf/cvpr/DongLPS0HL18,DBLP:conf/cvpr/XieZZBWRY19}, we use an ensemble training strategy with input diversity enhancement to encourage $G$ to generate adversarial examples with high transferability and black-box attack success rate. The adversarial is defined as:

\begin{equation}\label{eq:LossADV_G_ED}
\begin{aligned}
    L^{adv}_G &= \frac{1}{2K}\sum^K_{k=1}1 - \cos [M_k(z),M_k(T(G(x,y), p))]\\
    &+ \frac{1}{2K}\sum^K_{k=1}1 - \cos [M_k(z),M_k(T(G(y,x), p))],
\end{aligned}
\end{equation}
where $M_k$ represents the feature extractor of the $k$-th local pre-trained white-box model, and we use cosine similarity as the distance function. $T(\cdot)$ represents the transformation function, and $p$ is a pre-defined probability of whether the transformation will be conducted upon $G(x,y)$. Specifically, we choose image resizing and Gaussian noising as the transformation function. Both of them can degrade the attack strength of adversarial examples whose adversarial modifications have faint transferability among different black-box models. 

Accordingly, the adversarial attack loss of $H$ is defined as: 

\begin{equation}\label{eq:LossADV_G_ED}
\begin{aligned}
   L^{adv}_H &= \frac{1}{2K}\sum^K_{k=1}1 - \cos [M_k(x),M_k(H(G(x,y)))]\\
   &+ \frac{1}{2K}\sum^K_{k=1}1 - \cos [M_k(y),M_k(H(G(y,x)))],
\end{aligned}
\end{equation}
note that the input diversity is not included as it is unnecessary for $H$ to own transferability.

\input{section/algorithm}

\textbf{Auxiliary Objectives.} The histogram matching~\cite{DBLP:conf/mm/LiQDLYZL18}, denotes as $HM(x,y)$, is usually used to simulate the color distribution of reference $y$ while preserves the content information of $x$. Here, we use this objective function to ensure the makeup similarity on lips, eye shadows, and face regions as well as the reconstruction ability of $H$. In detail, the makeup loss is defined as:

\begin{equation}\label{eq:LossMakeup_G}
\begin{aligned}
    L^{make}_G &= \parallel G(x,y) - HM(x,y) \parallel_2
    \\&+\parallel G(y,x) - HM(y,x) \parallel_2,
\end{aligned}
\end{equation}
\begin{equation}\label{eq:LossMakeup_H}
\begin{aligned}
    L^{make}_H &= \parallel H(G(x,y)) - HM(x,y) \parallel_2
    \\&+ \parallel H(G(y,x)) - HM(y,x) \parallel_2.
\end{aligned}
\end{equation}


In addition, the generator $G$ and the regularization module $H$ are expected to preserve the original content and style information when the reference image is the source image itself, \black{which is called   self-reconstruction. This objective is significantly important for the generator \textit{G} to maintain the structure information of resource images and avoid distortion of face attributes.} The self-reconstruction path is defined as:

\begin{equation}\label{eq:Lossidt_H}
\begin{aligned}
    L^{idt}_{G,H} &= \parallel H(G(x,x))  -x\parallel_1 + LPIPS(H(G(x,x)), x)
    \\&+ \parallel  H(G(y,y))  -y\parallel_1 + LPIPS(H(G(y,y)), y),
\end{aligned}
\end{equation}
where the \textit{LPIPS}~\cite{DBLP:conf/cvpr/ZhangIESW18} function measures perceptual similarity  between two images.





\textbf{Total Loss.} The total loss for $D_X$ and $D_Y$ is as follow:
\begin{equation}\label{eq:LossD}
\begin{aligned}
    L_D = L^{gan}_D\Lambda^\mathrm{T}.
\end{aligned}
\end{equation}

The total loss of $G$ is defined as:
\begin{equation}\label{eq:LossG}
\begin{aligned}
    L_G = (L^{gan}_G, L^{reg}_G, L^{adv}_G, L^{make}_G, L^{idt}_{G,H})\Lambda^\mathrm{T},
\end{aligned}
\end{equation}
and the total loss of $H$ is defined as:
\begin{equation}\label{eq:LossH}
\begin{aligned}
    L_H = (L^{gan}_H, L^{adv}_H, L^{make}_H, L^{idt}_{G,H})\Lambda^\mathrm{T},
\end{aligned}
\end{equation}
where $\Lambda = (\lambda_{gan}, \lambda_{reg}, \lambda_{adv}, \lambda_{make}, \lambda_{idt})$ represents the hyper-parameters. The entire training process is illustrated in Alg.~\ref{alg}.

%% file: section/algorithm.tex
\begin{algorithm}[t]
\caption{The complete training process of AMT-GAN}
\label{alg}
\textbf{Input}: Source image  set $X$; reference image set $Y$; target image $z$; generator $G$; regularization module $H$; discriminators $D_X, D_Y$; local models $M$; optimizer $Adam$.\\
\textbf{Parameter}: Iterations $T$; hyper-parameters $\Lambda$.\\
\textbf{Output}: parameters $\omega_G,\omega_{D_X},\omega_{D_Y},\omega_H$ for networks $D_X,D_Y,G,H$.
\begin{algorithmic}[1] 
\STATE Initialize $\omega_G,\omega_{D_X},\omega_{D_Y},\omega_H$.
\FOR{$i = 0$ to $T-1$}
\STATE Randomly select source image $x \in X$ and reference image $y \in Y$ as the input of generator $G$;
\STATE \textbf{Updating $D_X$ and $D_Y$ with fixed $G$ and $H$};
\STATE Calculate $L_D$ in Eq. (\ref{eq:LossD});
\STATE $\omega_{D_X} \gets Adam(\omega_{D_X} ,L_D)$;
\STATE $\omega_{D_Y} \gets Adam(\omega_{D_Y} ,L_D)$;
\STATE \textbf{Updating $G$ with fixed $D$ and $H$};
\STATE Calculate $L_G$ in Eq. (\ref{eq:LossG});
\STATE $\omega_{G} \gets Adam(\omega_{G} ,L_G)$;
\STATE \textbf{Updating $H$ with fixed $G$ and $D$};
\STATE Calculate $L_H$ in Eq. (\ref{eq:LossH});
\STATE $\omega_{H} \gets Adam(\omega_{H} ,L_H)$;
\ENDFOR
\STATE \textbf{return} $\omega_G,\omega_{D_X},\omega_{D_Y},\omega_H$.
\end{algorithmic}
\end{algorithm}


%% file: section/4-experiments.tex
\section{Experiments}\label{experiemnt}
\begin{figure*}[t]
\setlength{\belowcaptionskip}{-0.3cm}
\centering
\includegraphics[width=0.85\textwidth]{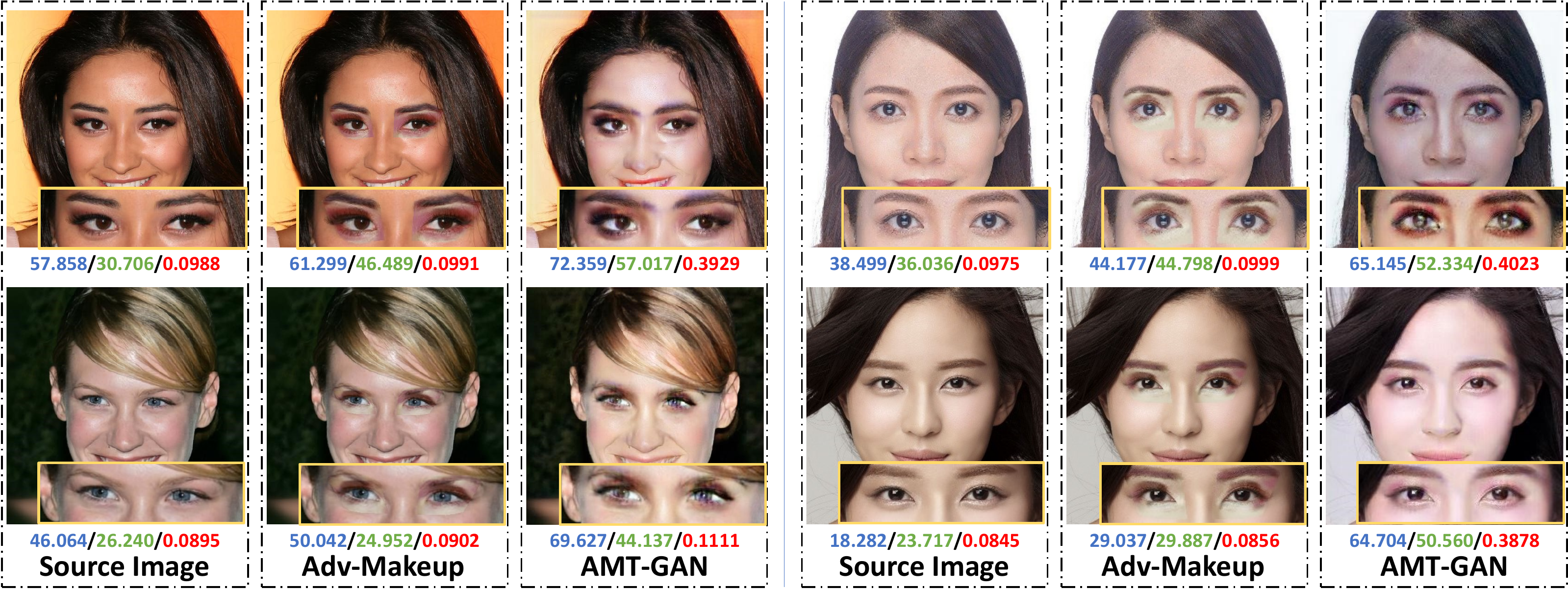}
\caption{Comparison of visual quality between Adv-Makeup and AMT-GAN. 
The numbers under each image stand for the confidence scores returned from commercial APIs.}

\label{fig.compare_ijcai}
\end{figure*}

\begin{table*}
\setlength{\belowcaptionskip}{-0.3cm}
\centering
\resizebox{0.85\textwidth}{!}
{
\begin{tabular}{cl|cccc|cccc}
\hline
\multicolumn{1}{l}{} & \multicolumn{1}{l|}{} & \multicolumn{4}{c|}{CelebA-HQ}        & \multicolumn{4}{c}{LADN-dataset}      \\ \hline
\multicolumn{1}{l}{} &                       & IRSE50 & IR152 & Facenet & Mobileface & IRSE50  & IR152  & Facenet  & Mobileface \\ \hline
\multicolumn{2}{c|}{\black{Clean}}               & 7.29 & 3.80 & 1.08 & 12.68 & 2.71 & 3.61 & 0.60 & 5.11 \\ \hline
\multicolumn{2}{c|}{PGD~\cite{DBLP:conf/iclr/MadryMSTV18}}                      & 36.87  & 20.68 & 1.85    & 43.99      & 40.09   & 19.59  & 3.82     & 41.09      \\
\multicolumn{2}{c|}{MI-FGSM~\cite{DBLP:conf/cvpr/DongLPS0HL18}}                  & 45.79  & 25.03 & 2.58     & 45.85      & 48.9    & 25.57  & 6.31     & 45.01     \\
\multicolumn{2}{c|}{TI-DIM~\cite{DBLP:conf/cvpr/DongPSZ19}}                    & \cellcolor[rgb]{ .973,  .796,  .678}63.63  & \cellcolor[rgb]{ .973,  .796,  .678}36.17 & \cellcolor[rgb]{ .988,  .894,  .839}15.3   & \cellcolor[rgb]{ .957,  .69,  .518}57.12      & \cellcolor[rgb]{ .988,  .894,  .839}56.36   & \cellcolor[rgb]{ .988,  .894,  .839}34.18  & \cellcolor[rgb]{ .988,  .894,  .839}22.11    & \cellcolor[rgb]{ .973,  .796,  .678}48.30        \\
\multicolumn{2}{c|}{Adv-Makeup~\cite{DBLP:conf/ijcai/YinWYGKDLL21}}               & 21.95  & 9.48  & 1.37    & 22.00      & 29.64   & 10.03  & 0.97     & 22.38      \\ 
\multicolumn{2}{c|}{\black{TIP-IM}~\cite{yang2021towards}}               & \cellcolor[rgb]{ .988,  .894,  .839}54.4  & \cellcolor[rgb]{ .957,  .69,  .518}37.23  & \cellcolor[rgb]{ .957,  .69,  .518}40.74  & \cellcolor[rgb]{ .988,  .894,  .839}48.72  & \cellcolor[rgb]{ .973,  .796,  .678}65.89 & \cellcolor[rgb]{ .973,  .796,  .678}43.57  & \cellcolor[rgb]{ .957,  .69,  .518}63.50  & \cellcolor[rgb]{ .988,  .894,  .839}46.48         \\
\hline
\multicolumn{2}{c|}{AMT-GAN}              & \cellcolor[rgb]{ .957,  .69,  .518}76.96  & \cellcolor[rgb]{ .988,  .894,  .839}35.13 & \cellcolor[rgb]{ .973,  .796,  .678}16.62   & \cellcolor[rgb]{ .973,  .796,  .678}50.71      & \cellcolor[rgb]{ .957,  .69,  .518}89.64   & \cellcolor[rgb]{ .957,  .69,  .518}49.12 & \cellcolor[rgb]{ .973,  .796,  .678}32.13    & \cellcolor[rgb]{ .957,  .69,  .518}72.43      \\ \hline
\end{tabular}
}
\caption{Evaluations of \textit{ attack success rate} (ASR) for black-box attacks}
\label{Tab.result}
\end{table*}

\subsection{Experimental Setting}
\textbf{Implementation details. } We construct the architecture of $G$, $D_X$, and $D_Y$ in AMT-GAN following~\cite{DBLP:conf/cvpr/JiangLG0HFY20}. For the training process, the hyper-parameters $\lambda_{GAN}$, $\lambda_{reg}$, $\lambda_{adv}$, $\lambda_{make}$, and $\lambda_{idt}$ are set to be $10$, $10$, $5$, $2$, and $5$ respectively. We train the AMT-GAN by an Adam optimizer~\cite{DBLP:journals/corr/KingmaB14} with the learning rate of $0.0002$, and set exponential decay rates as $(\beta_1, \beta_2) = (0.5, 0.999)$.

\textbf{Competitors.} We implement multiple benchmark schemes of adversarial attack, including PGD~\cite{DBLP:conf/iclr/MadryMSTV18}, MI-FGSM~\cite{DBLP:conf/cvpr/DongLPS0HL18}, TI-DIM~\cite{DBLP:conf/cvpr/DongPSZ19}, \black{TIP-IM~\cite{yang2021towards}}, and Adv-Makeup~\cite{DBLP:conf/ijcai/YinWYGKDLL21}, to serve as the competitors for comparison. Note that PGD, MI-FGSM, and TI-DIM are very famous for their strong attack ability, \black{TIP-IM is a very recent work which leverages adversarial examples to protect facial privacy}, and Adv-makeup is the most relative scheme to ours which also exploits the makeup transfer to generate adversarial face images with transferability. 

\textbf{Datasets. }Following~\cite{DBLP:conf/cvpr/ChenHWTSC19,DBLP:conf/cvpr/JiangLG0HFY20}, the \textit{Makeup Transfer} (MT) dataset~\cite{DBLP:conf/mm/LiQDLYZL18} is used as the training dataset, which consists of $1115$ non-makeup images and $2719$ makeup images. We choose two datasets as our test sets: (1) CelebA-HQ~\cite{karras2017progressive} is a widely used face image dataset with high quality.
For the testing, we select a subset of CelebA-HQ, which contains $1,000$ face images with different identities. (2) LADN-dataset~\cite{DBLP:conf/iccv/GuWCTT19} is a makeup dataset which contains $333$ non-makeup images and $302$ makeup images. We use $332$ non-makeup images as the test images. We divide all the test images into $4$ groups and aim images in each group to the same target identity.

\textbf{Target models. }Following~\cite{DBLP:conf/ijcai/YinWYGKDLL21}, we conduct extensive experiments to attack  $4$ popular black-box FR models, which include IR152~\cite{DBLP:conf/cvpr/HeZRS16}, IRSE50~\cite{DBLP:conf/cvpr/HuSS18}, Facenet~\cite{DBLP:conf/cvpr/SchroffKP15}, and Mobileface~\cite{DBLP:conf/cvpr/DengGXZ19}, and  $3$  commercial FR APIs including Face++, Aliyun, and Microsoft Azure. 

\begin{figure*}[t] 
\setlength{\belowcaptionskip}{-0.3cm}
    \centering
    \begin{subfigure}{0.245\textwidth}
        \centering
        \includegraphics[width=\textwidth]{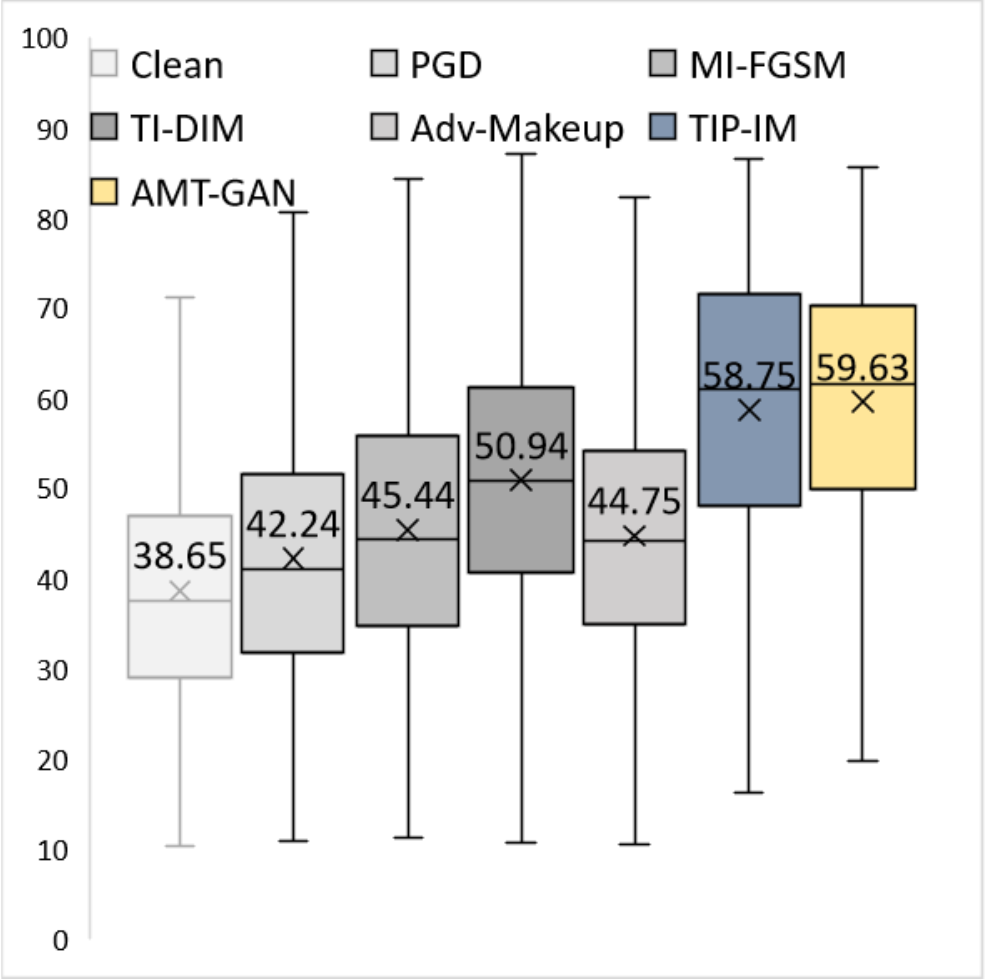} 
        \caption{CelebA-HQ on Face++}
        \label{Fig.face++HQ}
    \end{subfigure}
    \hfill
    \begin{subfigure}{0.245\textwidth}
        \centering
        \includegraphics[width=\textwidth]{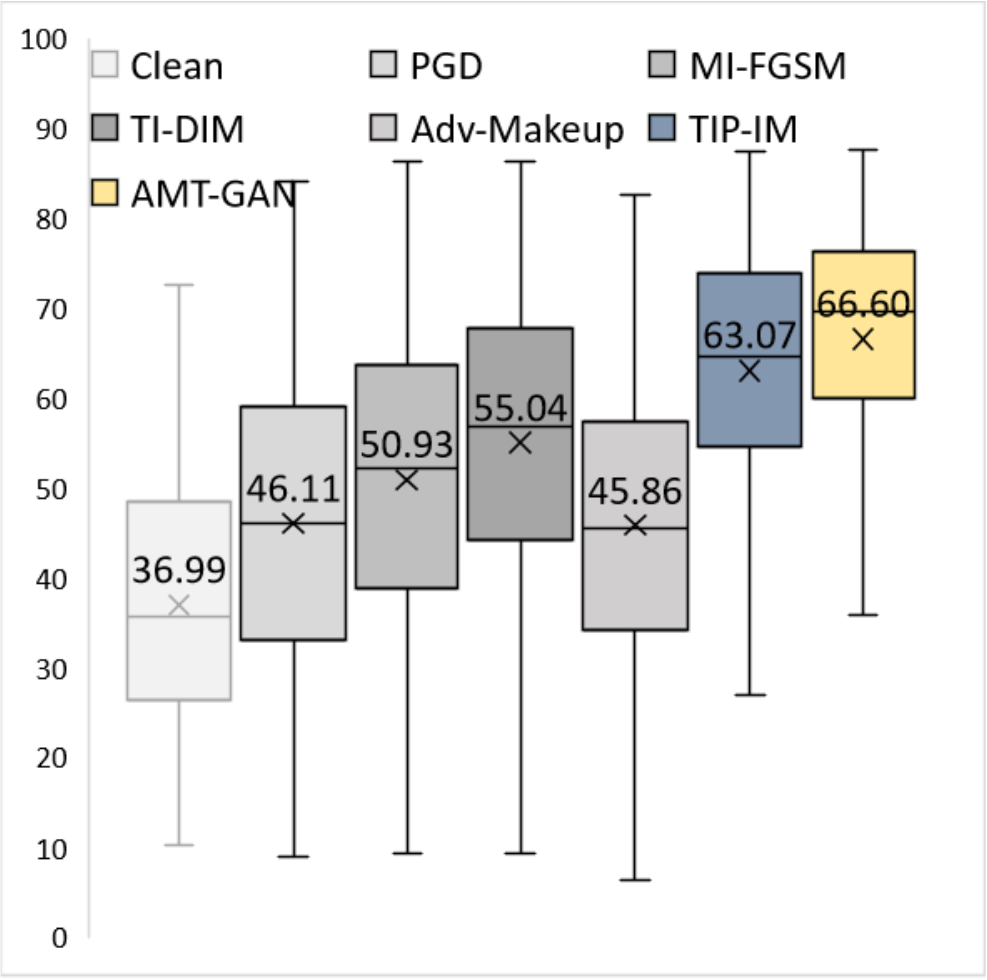}
        \caption{LADN-dataset on Face++}
        \label{Fig.face++LADN}
    \end{subfigure}
    \hfill
    \begin{subfigure}{0.245\textwidth}
        \centering
        \includegraphics[width=\textwidth]{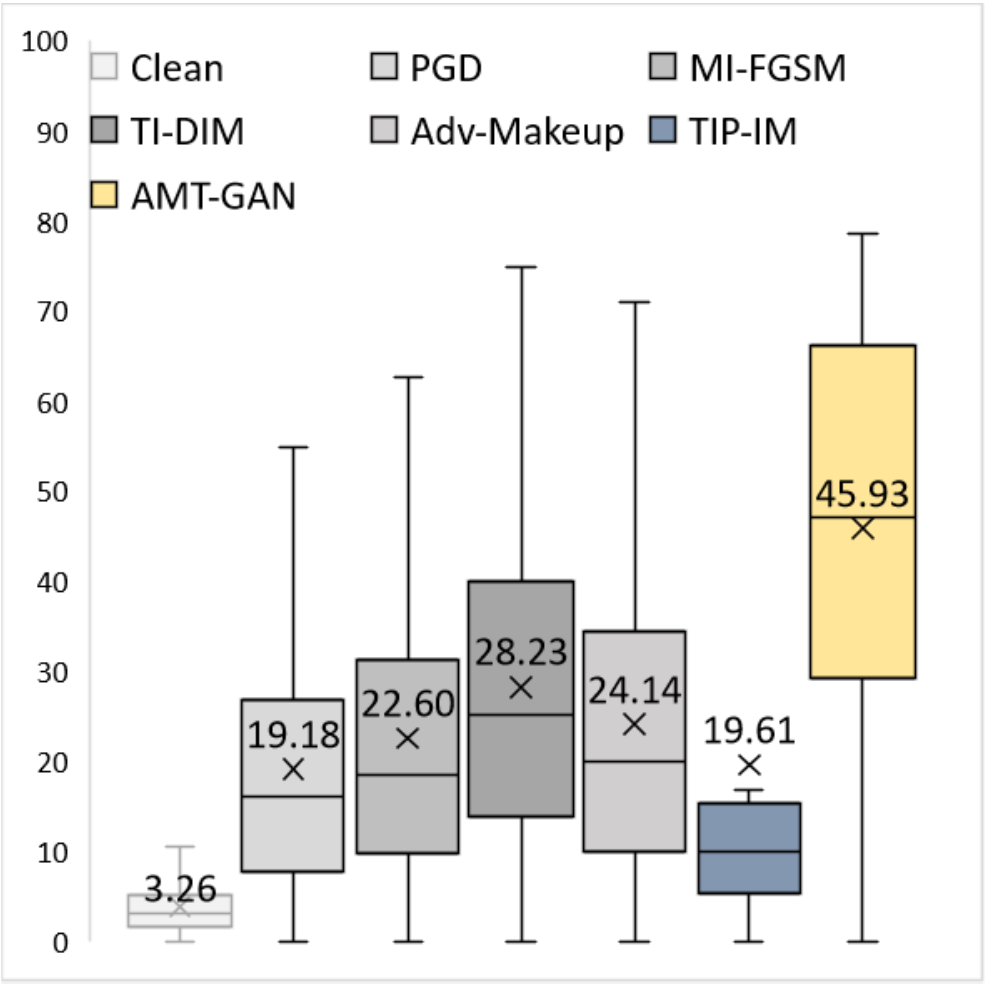} 
        \caption{CelebA-HQ on Aliyun}
        \label{Fig.aliyun-HQ}
    \end{subfigure}
    \hfill
    \begin{subfigure}{0.245\textwidth}
        \centering
        \includegraphics[width=\textwidth]{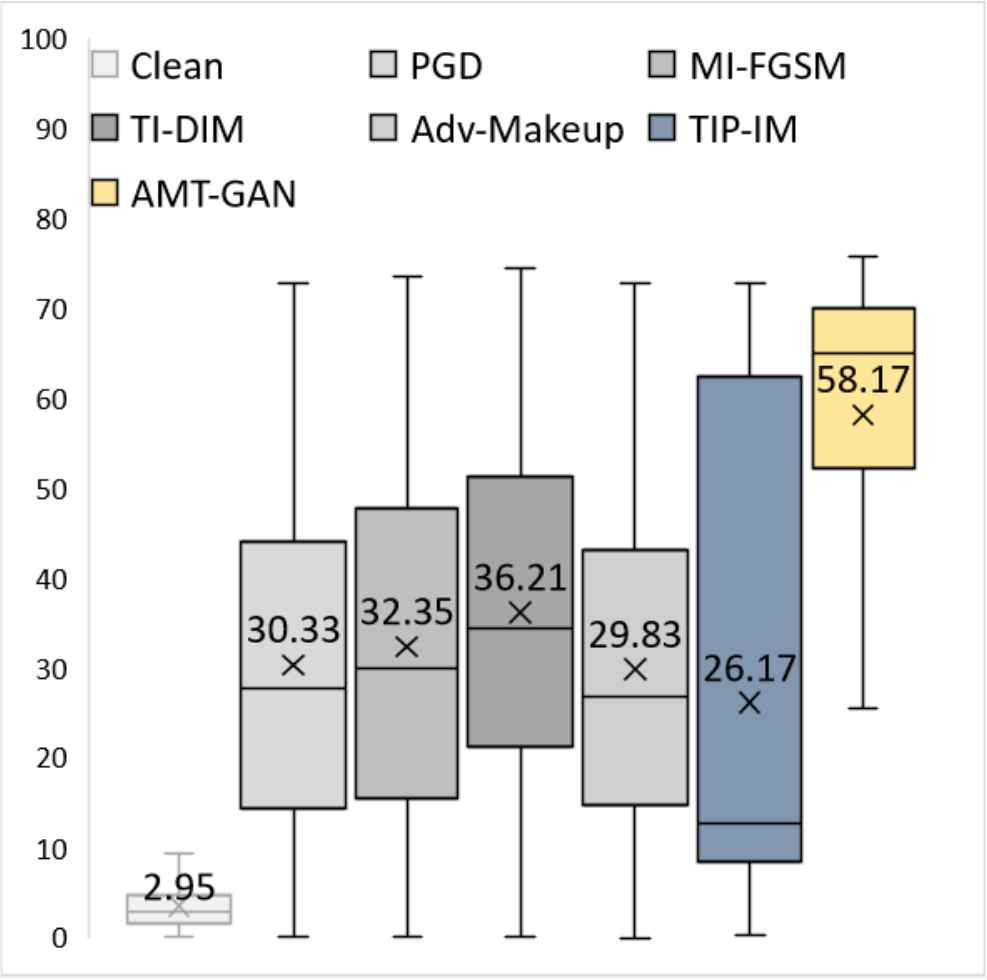} 
        \caption{LADN-dataset on Aliyun}
        \label{LFig.aliyun-LADN}
    \end{subfigure}
    \caption{Confidence scores returned from Face++ and Aliyun. \black{The mean confidence score of the runner-up (TIP-IM) for Face++ is $61.99$, while ours is $64.58$. For Aliyun, the mean confidence score  of the runner-up (TI-DIM) is $32.37$, while ours is $53.53$. We outperform the runner-ups about $4\%\sim60\%$.  Note that AMT-GAN also has a stronger transferability among different APIs while TIP-IM has a huge degradation in Aliyun compared with its performance on Face++.}}
    \label{Fig.APIs}
\end{figure*}

\textbf{Evaluation metrics. }Following existing impersonation attacks~\cite{DBLP:conf/ijcai/YinWYGKDLL21,DBLP:conf/cvpr/XiaoGFDGZ0021}, we use \textit{attack success rate} (ASR) to evaluate the attack ability of different methods. We calculate the ASR at FAR@$0.01$ for black-box testing. For commercial APIs, we directly record the confidence scores returned by FR servers. A higher confidence score represents that the victim FR API believes the two input images are of the same person with a higher probability. \black{We also leverage FID~\cite{DBLP:conf/nips/HeuselRUNH17}, PSNR(dB), and SSIM~\cite{DBLP:journals/tip/WangBSS04} to evaluate the image quality. FID  measures the distance between two data distributions, which is often used to investigate whether a generated dataset is as natural as the dataset extracted from the real world. PSNR and SSIM are widely-used methods to measure the difference between two images.}

\black{All of our experiments are conducted on RTX3090 GPU 24GB$*1$. For more experimental results, please refer to our supplementary. }

\subsection{Comparison Study}

\textbf{Evaluations on black-box attacks. }
Tab.~\ref{Tab.result} shows black-box attacks on four different pre-trained models which have high accuracy on public datasets. For each target model, the other three models will serve as the ensemble training model. Note that for Adv-makeup these three models will serve as the meta-learning model. The results show that AMT-GAN has a strong attack ability in the black-box setting.

\textbf{\black{Evaluations on image quality. }}
\black{Tab.~\ref{Tab.quality}   shows the quantitative evaluations on image quality. Notably, compared with TIP-IM, although our method have a worse performance in terms of PSNR(db) and SSIM, AMT-GAN performs better for the FID result.
This shows that the images generated by our method have more natural appearances than TIP-IM, although they get more information changed. This verifies our insight that arranging perturbations, instead of simply restricting them, is more important. We further attach 
the results of PSGAN~\cite{DBLP:conf/cvpr/JiangLG0HFY20}, which is the most famous makeup transfer scheme, to show that it is normal  to obtain similar evaluation results with these three metrics in the field of makeup transfer.}

\black{In addition, Adv-makeup seems to behave well in all the quantitative evaluations. However, it has an extremely low attack success rate as demonstrated in Tab.~\ref{Tab.result}. Furthermore, we give 
a qualitative comparison of visual image quality between Adv-Makeup and AMT-GAN, both of which  construct  adversarial face images based on makeup transfer. As shown in Fig.~\ref{fig.compare_ijcai}, the images generated by Adv-Makeup have sharp margins among the eyes region. On the contrary, AMT-GAN has a more realistic makeup style with smooth details. This is because Adv-makeup only changes the eyes region of original faces in a patch-based way, which leads to a good performance on quantitative evaluations, but leaves insufficient black-box attack strength and unresolved margin problem. 
}


\subsection{Attack Performance on Commercial APIs}  Fig.~\ref{Fig.APIs} illustrates the attack performance towards Aliyun and Face++  for each test dataset. We collect and average the  confidence scores from these APIs with massive adversarial examples. The results show that for both APIs, AMT-GAN outperforms competitors with regards to the attack ability.

\begin{table}[h]
\setlength{\belowcaptionskip}{-0.5cm}
\centering
\resizebox{0.4\textwidth}{!}
{
    \begin{tabular}{c|ccc}
    \hline
          & FID($\downarrow$)   & PSNR($\uparrow$)  & SSIM($\uparrow$) \\
    \hline
    Adv-Makeup~\cite{DBLP:conf/ijcai/YinWYGKDLL21} & 4.2282  & 34.5152  & 0.9850  \\
    TIP-IM~\cite{DBLP:conf/cvpr/DongPSZ19} & 38.7357  & 33.2089  & 0.9214  \\
    PSGAN~\cite{DBLP:conf/cvpr/JiangLG0HFY20} & 27.6765  & 18.1403  & 0.8041  \\
    AMT-GAN  (w/oH)  & 37.5486  & 19.3132  & 0.7807  \\
    AMT-GAN & 34.4405  & 19.5045  & 0.7873  \\
    \hline
    \end{tabular}%
}
    \caption{\black{Quantitative evaluations of image quality. AMT-GAN (w/oH) represents the AMT-GAN trained without the regularization module.}}
  \label{tab:addlabel}%
\label{Tab.quality}
\end{table}%

\subsection{Ablation Studies}
\begin{figure}[t]
\setlength{\belowcaptionskip}{-0.5cm}
\centering
\includegraphics[width=0.45\textwidth]{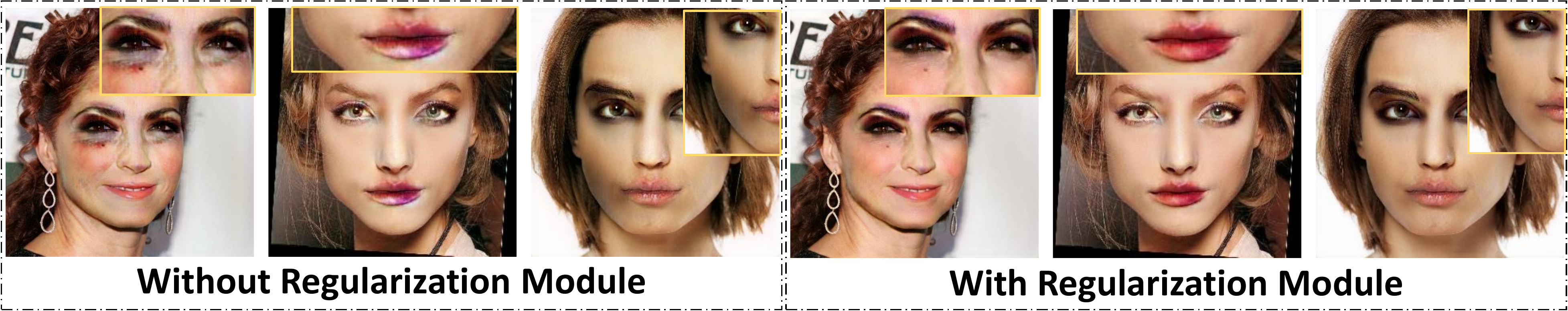}
\caption{Ablation study for the regularization module. The generator trained without the regularization module will generate images with unnatural details. Quantitative results are in Tab.~\ref{Tab.quality}.}
\label{fig.ablation_H}
\end{figure}

\textbf{Regularization module.} Here we  demonstrate the importance of the regularization module in maintaining the effectiveness and stability of AMT-GAN. As shown in Fig.~\ref{fig.ablation_H} and Tab.~\ref{Tab.quality}, in the absence of the regularization module, the generator is likely to generate images \black{with worse image quality}, which indicates that the mappings between style domains are damaged to some degree. We owe this to the fact that  the adversarial toxicity has poisoned the cycle reconstruction path. By applying the regularization module,  the regularized cycle consistency loss can make the outputs of the generator more natural.

\begin{figure}[ht] 
\setlength{\belowcaptionskip}{-0.5cm}
\centering
\begin{subfigure}{0.45\textwidth}
    \centering
    \includegraphics[width=\textwidth]{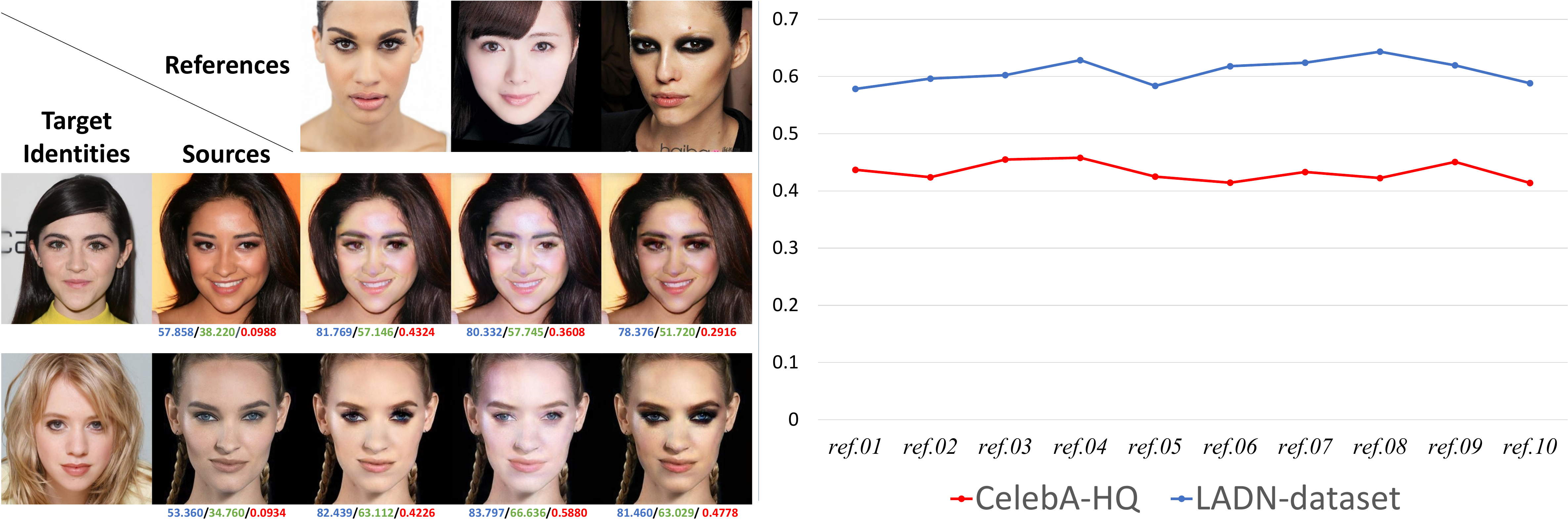} 
\end{subfigure}
\caption{Evaluating the impact of different makeup styles. The left columns are adversarial examples under different references.The right figure illustrates average ASR towards different black-box models in Tab.~\ref{Tab.result} with $10$ different references. }
\label{Fig.style_gener}
\end{figure}

\textbf{Style-robust makeup transfer.} It is commonly expected that we can generate adversarial images with satisfied visual quality for any given makeup style. Thus it is desired to evaluate the impact of different references. 
We randomly choose $10$ images from MT-dataset and LADN-dataset with different makeup styles as the references for testing.
As illustrated in Fig.~\ref{Fig.style_gener}, AMT-GAN is robust to the changes of makeup style, where the targeted adversarial face images maintain a good balance between the content of source images and the makeup style of references. 
The right figure of Fig.~\ref{Fig.style_gener} shows that the changes of references have weak impact on the attack strength.


%% file: section/6-Limitation.tex
\section{Limitations and Future Work}\label{Limitation}
Although AMT-GAN shows effectiveness on attacking commercial APIs, it tends to have a higher attack strength and a better visual quality in images of female, which is caused by the unbalance of gender in makeup transfer training dataset (\eg, MT-dataset~\cite{DBLP:conf/mm/LiQDLYZL18}). We believe that this problem can be solved by  developing a 
more general and comprehensive training dataset. 

Another problem with AMT-GAN is that the structure information sometimes gets slightly unaligned although we have designed corresponding objective functions for alleviation. The same problem also exists in the field of makeup transfer~\cite{DBLP:conf/cvpr/JiangLG0HFY20,DBLP:conf/iccv/GuWCTT19}, and may become worse in generating adversarial examples. We leave this point to our future works. 


Finally, the experimental results in Tab.~\ref{Tab.quality}  show that existing popular metrics used to evaluate the quality of images are unsuitable for the scenario of makeup transfer. New metrics are needed to evaluate whether a face image appears more natural than another one.
Besides, it is still desirable for us to further improve the visual quality of adversarial faces.  It is amazing to restrict the regions of makeup transfer to a small area (\eg,  eyes), which can also preserve a high attack success rate.  
We also leave these to our future works.

%% file: section/5-conclusion.tex
\section{Conclusion}\label{Conclusion}
In this paper, focusing on protecting facial privacy against malicious deep \textit{face recognition} (FR) models, we propose AMT-GAN to construct adversarial examples that achieve a stronger attack ability in the black-box setting, while maintaining a better visual quality. AMT-GAN is able to generate adversarial face images with makeup transferred from any reference image. The experiments over multiple datasets and target models show that AMT-GAN is highly effective towards different  open-source FR models and commercial APIs, and achieves a satisfied balance between the visual quality of adversarial face images and their attack strength.

%% file: section/Appendix.tex
\begin{figure*}[t]
\setlength{\belowcaptionskip}{-0.5cm}
\centering
\includegraphics[width=1\textwidth]{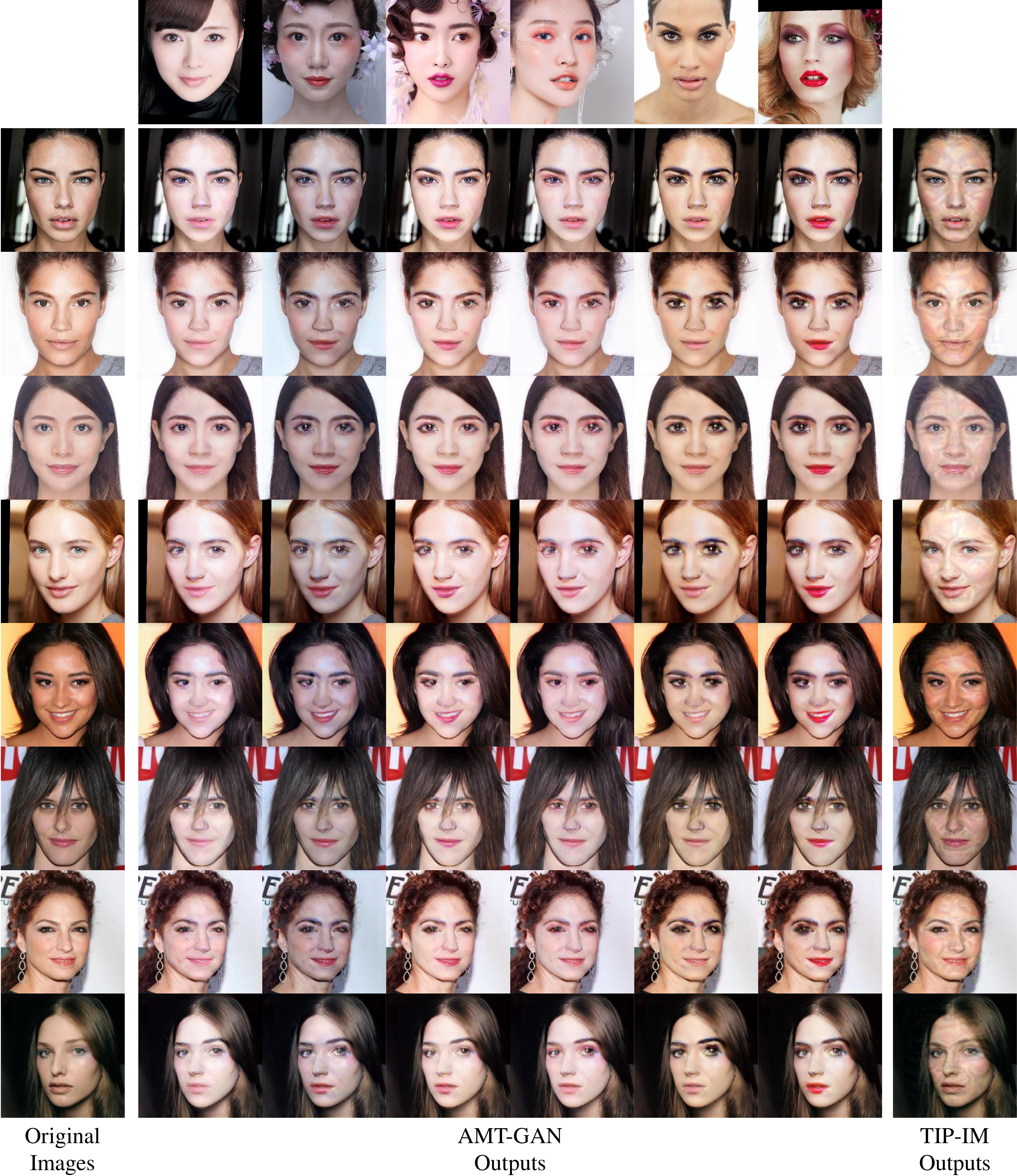}
\caption{Results from AMT-GAN and TIP-IM. Here we show the images from the same outputs dataset which we use for evaluations in our paper. The images on the top row are the cosmetic references for makeup transfer. Please zoom in for a better view.}
\label{fig:samples}
\end{figure*}

\begin{figure*}[t]
\setlength{\belowcaptionskip}{-0.5cm}
\centering
\includegraphics[width=1\textwidth]{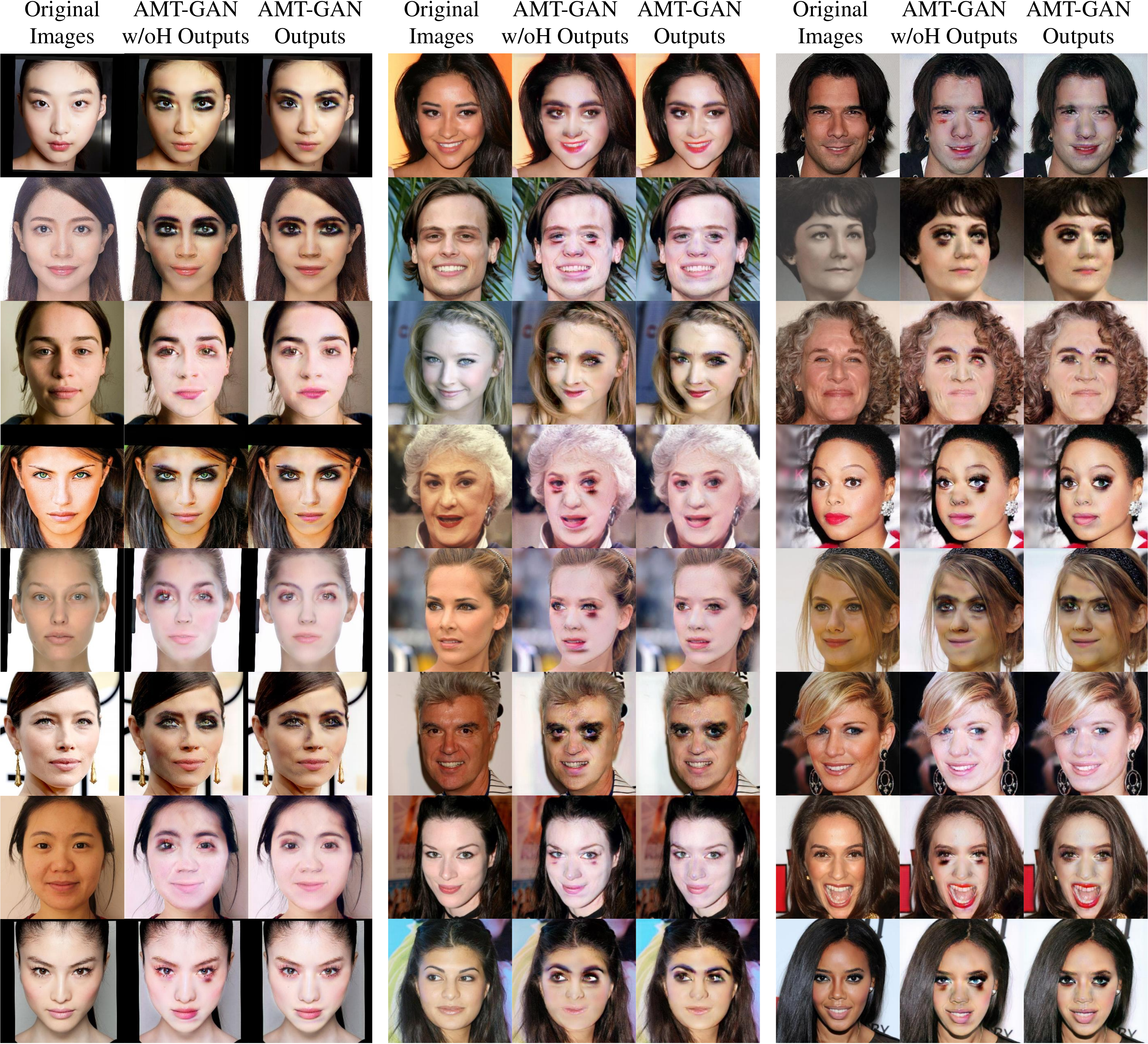}
\caption{Visual results of ablation study about regularization module. The images from the generator trained without the regularization module suffer from fake shadows, distortion of structure information, unaligned makeup position, etc., which are typical indications of weak domain mappings, caused by damaged cycle consistency loop by adversarial toxicity in the training phase. As the visual results here and the quantitative results in our paper have shown, the regularization module can eliminate or alleviate this phenomenon. Notably, as makeup transfer is still in development and may have some little issues, 
a small minority of images (no matter with or without regularization module) may have asymmetrical eye-shadow, which is beyond the scope of our investigation in this paper. Please zoom in for a better view.} 
\label{fig:woh_samples}
\end{figure*}